\documentclass{article}

\usepackage{PRIMEarxiv}
\usepackage{graphicx}
\usepackage{amsmath}
\usepackage{tikz}
\usetikzlibrary{decorations.pathmorphing,patterns,shapes,arrows,positioning,calc}
\tikzstyle{block} = [rectangle, draw, fill=white, line width=0.5mm,
    text width=5em, text centered, rounded corners, minimum height=4em]
\usepackage[utf8]{inputenc} 
\usepackage[T1]{fontenc}    
\usepackage{hyperref}       
\usepackage{url}            
\usepackage{booktabs}       
\usepackage{amsfonts}       
\usepackage{nicefrac}       
\usepackage{microtype}      
\usepackage{lipsum}
\usepackage{fancyhdr}       
\usepackage{graphicx}       
\graphicspath{{media/}}     

\title{Graph Neural Networks for Dynamic Modeling of Roller Bearing
}

\author{
  Vinay Sharma\\
  Intelligent Maintenance and Operations Systems \\
  EPFL \\
  Lausanne, Switzerland\\
  \texttt {vinay.sharma@epfl.ch} \\
   \And
  Jens Ravesloot \\
  SKF, Research and Technology Development \\
  Houten, the Netherlands\\
  \texttt{jens.ravesloot@skf.com} \\
  \And
  Cees Taal \\
  SKF, Research and Technology Development \\
  Houten, the Netherlands\\
  \texttt{cees.taal@skf.com} \\  
  \And
  Olga Fink \\
  Intelligent Maintenance and Operations Systems \\
  EPFL \\
  Lausanne, Switzerland\\
  \texttt{olga.fink@epfl.ch} \\  
  }

\begin{document}
\maketitle

\begin{abstract}
In the presented work, we propose to apply the framework of graph neural networks (GNNs) to predict the dynamics of a rolling element bearing. This approach offers generalizability and interpretability, having the potential for scalable use in real-time operational digital twin systems for monitoring the health state of rotating machines. By representing the bearing's components as nodes in a graph, the GNN can effectively model the complex relationships and interactions among them. We utilize a dynamic spring-mass-damper model of a bearing to generate the training data for the GNN. In this model, discrete masses represent bearing components such as rolling elements, inner raceways, and outer raceways, while a Hertzian contact model is employed to calculate the forces between these components.

We evaluate the learning and generalization capabilities of the proposed GNN framework by testing different bearing configurations that deviate from the training configurations. Through this approach, we demonstrate the effectiveness of the GNN-based method in accurately predicting the dynamics of rolling element bearings, highlighting its potential for real-time health monitoring of rotating machinery.
\end{abstract}


\section{Introduction}
\label{sec:intro} 
Real-time condition monitoring is essential for realizing real-time operational digital twins of complex systems such as rotating equipment. Digital twins enable real-time fault diagnosis and prognosis, mitigating the risk of catastrophic system failures and reducing maintenance costs through early intervention in case of faults. However, purely data-driven methods often struggle to capture the underlying dynamics and generalize to operating conditions not included in the training datasets. Consequently, they fall short of accurately predicting the long-term evolution of physical system states. 

To address these challenges, physics-informed neural networks (PINNs) have emerged as potential solution. PINNs integrate the partial differential equation (PDE) of the underlying system into the loss function, thereby regularizing the solution learned by the neural network. These methods have demonstrated significant success in various mechanics problems, including stress prediction in homogeneous elastic plates~\cite{Hag_ehsan}, composites~\cite{Yan}, and heterogeneous materials~\cite{PINN_alex}. In the context of multibody dynamical systems (MBD), the PINN loss can be formulated based on the Lagrangian of the system~\cite{Lutter}, Hamiltonian~\cite{greydanus}, or conservation of energy~\cite{jagtapCpinn}.

However, applying PINNs to systems with a large number of components presents challenges. It requires explicit derivation of either the PDE or analytical expressions for the conserved quantities, which can be cumbersome for complex multi-component systems. Additionally, enforcing boundary conditions becomes challenging, particularly in multi-component systems where boundaries dynamically form due to contact between different components~\cite{Gao_PINN}. Therefore, to handle a large number of interacting components, a network with an encoded inductive bias in its architecture is necessary.

Graph neural networks (GNNs)~\cite{battaglia, sanchez2020learning,pfaff2020} provide a promising solution to these challenges by representing input components as nodes in a graph and modeling interactions between them as messages passed over the edges of the graph. Due to their encoded inductive bias, they generalize well to systems with varying configurations and boundary conditions. A lot of physical systems consist of components that interact with each other which makes the graph structure of the GNNs very suitable.

In message-passing-GNNs (MP-GNNs), the topological structure of a multi-component system can be represented as a graph where nodes represent the state of different components and edges between the nodes represent the interactions between those components. The pairwise interactions are then modeled as messages passed over the edges. MP-GNNs comprise two networks:(i) an edge network that takes edge features between two nodes (e.g., the distance vector) and generates a message, and  (ii) a node network that takes the aggregated messages from all the neighboring nodes and produces a new node state. This process is repeated several times until the final node state is decoded as a target output. Depending on the task, the target of the graph neural network can be the predicted aggregated acceleration of the node.
These models have successfully been applied to various dynamics prediction tasks. They have shown success in simple systems such as particle and spring-mass systems~\cite{shlomi2020graph}, as well as in more complex scenarios like three-dimensional skeleton-based human motion prediction~\cite{Maosen}.

Expanding upon previous research on simple mass-spring systems, our study delves into the specificities of modelling bearing dynamics. Accurate fast modeling of bearing dynamics is vital for timely fault detection and failure prediction in rotating equipment. Building upon the efficacy of GNNs in capturing complex relationships and dynamics, our aim is to develop a graph-neural-network-based simulator that can accurately capture the complex interactions between different components in a bearing. Compared with Finite Element Analysis (FEA) or parameter-calibrated lumped parameter models, a GNN-based bearing model can have specific advantages. Specifically, a significant reduction in computational complexity \cite{pfaff2020}, where dynamics are learned solely from measurements without requiring knowledge of stiffness, mass and damping factors. Moreover, in contrast to pure data-driven methods, this method is interpretable (allowing for the inference of physical quantities not explicitly trained for), generalizable (enabling extrapolation to unseen conditions such as new shaft loads), and flexible (allowing the construction of new graphs, such as by changing the number of rolling elements in the bearing). 

In this work, we present the first proof of concept for the application of GNNs in modeling bearing dynamics. To achieve this, we train a GNN on a simple 2D dynamic bearing model, demonstrating its interpretability, generalizability, and flexibility. In future work, we anticipate that this concept can be extended to real sensor data and advanced FEA simulations, such as those involving complex elasto-hydrodynamic forces.
The proposed model incorporates specific node features, edge features, and graph connections that are essential for modelling a bearing as a graph. We further introduce the use of Message-Passing Graph Neural Networks (MP-GNNs) as proposed in \cite{sanchez2020learning} to predict the evolution of this dynamic. However, in contrast to   \cite{sanchez2020learning}, we propose modifications to the GNN architecture to decode roller loads from the edges and capture the dynamics of the raceways and rolling elements from the nodes.

This paper is organized as follows: In Section \ref{sec:bearing_model}, we first introduce a 2D dynamic bearing model. This analytical model serves as the generator of simulation trajectories on which we train the graph-based bearing model described in Section \ref{sec:GNN_model}. In Section \ref{sec:case_study}, we describe the bearing configurations and operating conditions used to generate the simulation data for training and testing the model. In Section \ref{sec:results}, we present the results of our experiments and evaluate the performance of the graph-based bearing model. Finally, in \ref{sec:conclusions}, we summarize the findings of the study, discuss their implications, and suggest future extensions of the proposed research study.

\section{Dynamic bearing model}
\label{sec:bearing_model} 
In this work, a 2D dynamic bearing model is used to simulate the behavior of a cylindrical roller bearing (CRB) for training and validation of the GNN. The chosen physics-based bearing model captures the essential components including the inner and outer rings, as well as multiple rolling elements \cite{Gupta1984} as depicted in figure \ref{fig:bearing_model_schematic_image}. Nonlinear contact models, based on the work of \cite{Lundberg1939} and \cite{palmgren1959}, are utilized to model the contacts between the rings and rolling elements. To establish the mechanical connections, the inner and outer rings are connected to the ground through springs and dampers. It is important to note that the bearing in this work is considered stationary and all bodies are restricted to horizontal and vertical movements only. The forces acting on each body are calculated as a function of their velocities and positions in space. Using the mass of the inner and outer ring, their respective accelerations can be computed and the Runge-Kutta method (RK4) is used to numerically integrate these to their updated positions and velocities. The rolling elements are assumed to have negligible mass and their positions are determined as a function of the inner and outer ring positions. 

To introduce external stimuli to the system, a  time-varying vertical force is applied to the outer ring as an input. The internal loads, positions, and velocities of all components serve as outputs and are used to train the GNN. Figure \ref{fig:bearing_model_schematic_image} shows schematic of the dynamic bearing model.
\begin{figure}[t]
    \centering  
    \begin{tikzpicture}[scale=0.4, every node/.style={transform shape},thick]
        \node(Center)[circle, draw,inner sep=48mm, rounded corners=8pt,ultra thick, fill=green!20!white] {};
        \node at (Center)[circle, draw,inner sep=43mm, rounded corners=8pt,ultra thick, fill=white] {};
        \filldraw (Center) [ ultra thick, fill=red!20!white] circle (20mm);
        \filldraw (Center) [ ultra thick, fill=white] circle (14mm);
        
        \foreach \x in {0,...,7} {
            \begin{scope}[black, shift={(0mm,0mm)},rotate=\x/8*360]
            \node[circle,fill=white,inner sep=6mm,draw=black] (rollingElement) at (0,39mm) {};
            \draw[decoration={aspect=0.5, segment length=1mm, amplitude=1.5mm,coil},decorate] (0,20mm) -- (rollingElement); 
            \draw[decoration={aspect=0.5, segment length=1.3mm, amplitude=1.5mm,coil},decorate] (0,60mm) -- (rollingElement); 
            \end{scope}
        };
        
        \draw (-10mm, 0) -- (10mm, 0);
        \fill [pattern=north east lines] (-10mm,0) arc (180:360:10mm);
        \draw[decoration={aspect=0.5, segment length=1.25mm, amplitude=1.5mm,coil},decorate] (-5mm, 13mm) -- (-5mm,0mm);
        \begin{scope}[xshift = 5mm, yshift = 0mm]
    		\draw (0,13mm) -- (0,7mm);
            \draw (0,6mm) -- (0,0);
            \draw (-2mm,6mm) -- (2mm,6mm);
            \draw (-3mm,7mm) -- (3mm,7mm);
            \draw (-3mm,7.3mm) -- (-3mm,4mm);
            \draw (3mm,7.3mm) -- (3mm,4mm);
    	\end{scope}
     
        \draw (-10mm, -80mm) -- (10mm, -80mm);
        \fill [pattern=north east lines] (-10mm, -80mm) arc (180:360:10mm);
        \begin{scope}[xshift = 0mm, yshift = -80mm]
    		\draw (0,12mm) -- (0,7mm);
            \draw (0,6mm) -- (0,0);
            \draw (-2mm,6mm) -- (2mm,6mm);
            \draw (-3mm,7mm) -- (3mm,7mm);
            \draw (-3mm,7.3mm) -- (-3mm,4mm);
            \draw (3mm,7.3mm) -- (3mm,4mm);
    	\end{scope}
        
    \end{tikzpicture}
    \caption{Schematic overview of the 2D bearing model used to generate signals for the GNN.}
    \label{fig:bearing_model_schematic_image}
\end{figure}
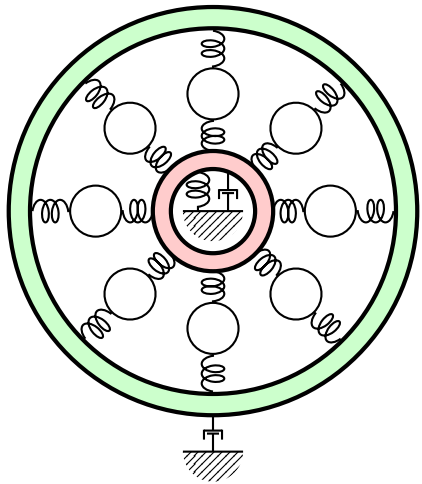
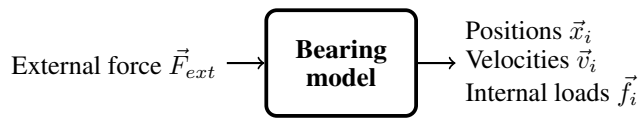
\begin{figure}[t]
    \centering  
    \begin{tikzpicture}[scale=1, every node/.style={transform shape},thick]
        \node [block] (model) {\bf Bearing model};
        \node [left=0.5cm of model, align=left] (input) {External force $\vec{F}_{ext}$};
        \node [right=0.5cm of model, align=left] (output) {Positions $\Vec{x}_i$\\Velocities $\Vec{v}_i$\\Internal loads $\vec{f}_{i}$};
        
        \draw[->] (input) -- (model);
        \draw[->] (model) -- (output);
    \end{tikzpicture}
    \caption{The inputs and outputs of the bearing model.}
    \label{fig:bearing_model_inputs_outputs_image}
\end{figure}


\section{A graph-based model of bearings}
\label{sec:GNN_model}
\begin{figure}
\centering
\includegraphics[scale=0.5,width=0.5\textwidth]{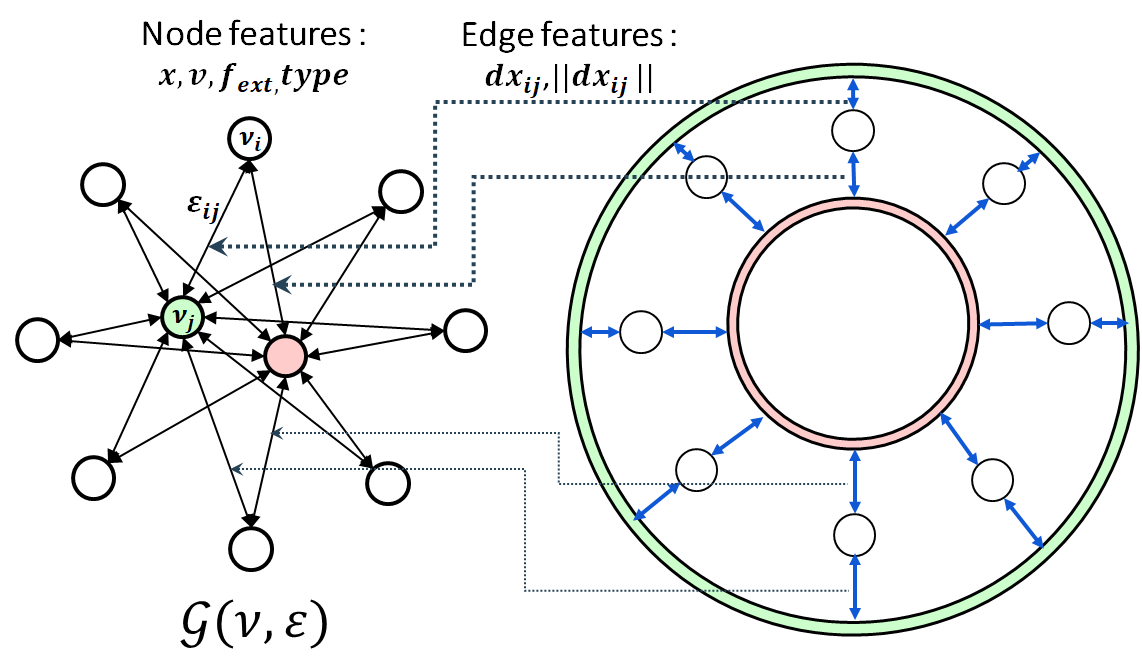}
\caption{Graph Representation of 2d Dynamic Model: The node features include the position $\Vec{x}_i$ and velocity $\Vec{v}_i$ of the centers of components (set to zero for rollers), external force $\Vec{F}_{ext}$, and node type. The edge features include the relative distance $\Vec{dx_{ij}}$ and its magnitude between the rollers and the circumferences of the inner and outer rings.}
\label{fig:graph_bearing}
\end{figure}
The bearing model can be effectively represented as a graph. Figure \ref{fig:graph_bearing} depicts a graph representation $\mathcal{G}={(\nu,\varepsilon})$ of the 2D dynamical model discussed in section \ref{sec:bearing_model}.
The graph representation captures the essential components of the model, including the inner ring, outer ring, and rolling elements, as nodes $\nu$ in the graph. The interactions between the rolling elements and the rings, characterized by non-linear contacts, are represented by edges $\varepsilon$  in the graph. 

In the following sections, we will elaborate on the GNN model, providing more details about the node and edge features, as well as the learning process.

\subsection{Node and edge features}
\textbf{Node Features}: The nodes in the graph represent the inner ring, outer ring, and rollers of the bearing system. Each node is characterized by a set of features denoted as $\nu_i={\Vec{x}_i,\Vec{v}_i,\Vec{F}_{ext},\text{type}}$. For the nodes representing the inner and outer rings, the feature $\Vec{x_i}$ corresponds to the position of their respective centers and their velocity is captured by the feature $\Vec{v_i}$, measured in millimeters per second.

For the nodes representing the rollers, the features $\Vec{x_i}$ and $\Vec{v_i}$  are set to zero. This choice reflects the assumption that the dynamics of rollers 
is purely governed by pair-wise interactions with the rings through relative positional features, which are encoded in the edges connecting the nodes.

In terms of external forces, the node representing the outer ring has a non-zero value for the feature $\Vec{F}_{ext}$, which accounts for the externally applied vertical radial force on the outer ring. Whereas, the inner ring and rollers have a value of zero for the external force feature $\Vec{F}_{ext}$, indicating that no external forces are applied to them.

To distinguish between the different components, the node type is encoded as a categorical variable. This allows for differentiation between the three types of components: the inner ring, outer ring, and rollers within the graph representation of the bearing system.
\\
\textbf{Edge Features}: The rollers are connected to the inner ring and outer ring nodes through bidirectional edges $\varepsilon_{ij} = {\Vec{dx_{ij}},||\Vec{dx_{ij}}||}$. These edges capture the 2D distance vector $\Vec{dx_{ij}}$ between the roller center and the circumference of the inner or outer ring, along with its scalar magnitude $||\Vec{dx_{ij}}||$.

The choice of using the distance vector and its magnitude as edge features is motivated by the assumption that the non-linear contact between the rollers and the rings can be modeled by non-linear springs. In this model, the forces depend solely on the elongation or compression of the springs, which is captured by the relative distance vector between the roller center and the ring circumference.

To calculate the 2D distance vector, the points on the circumferences of the inner and outer rings are determined, taking into account that the center of the roller is positioned midway between them. This approach ensures an accurate representation of the spatial relationship between the rollers and the rings, enabling the modeling of the contact forces and interactions within the bearing system.

It is worth highlighting that our approach distinguishes itself  from the previous applications of GNNs in predicting dynamics of spring-mass or particle systems \cite{sanchez2020learning, pfaff2020} by incorporating absolute position and velocity features on the nodes representing the inner and outer rings. While earlier applications primarily focused on capturing pair-wise interactions that are independent of position, our objective in modeling bearings expands to encompass the dynamics of both the inner and outer rings. This consideration takes into account  their interactions with the ground through springs and dampers as shown in Figure \ref{fig:bearing_model_schematic_image}.

\subsection{Model}
\begin{figure}
\centering
\includegraphics[scale=0.5,width=0.52\textwidth]{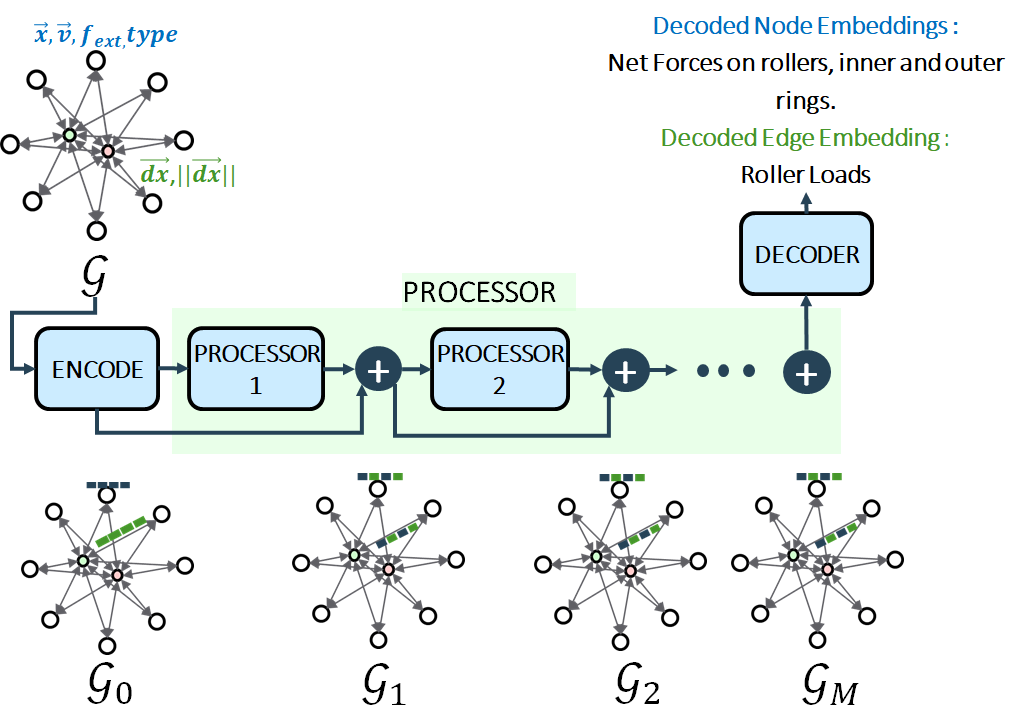}
\caption{Encode-Process-Decode architecture with message passing GNN: The Encoder transforms the graph $\mathcal{G}$ into $\mathcal{G}_0$.  Processor 1 takes $\mathcal{G}_0$ as input and transforms it  into $\mathcal{G}_1$ after a single message passing step. Subsequent Processors sequentially transform $\mathcal{G}_1$. Finally, the Decoder decodes both the latent nodes and edges of the graph $\mathcal{G}_M$.}
\label{fig:model}
\end{figure}

To predict the dynamics of the bearing, we utilize an encode-process-decode architecture, employing a message-passing graph neural network framework. The schematic of the model described in this section is illustrated in Figure \ref{fig:model}.

\textbf{Encode}: The encoder takes a graph $\mathcal{G}={(\nu,\varepsilon})$ and uses separate Multi-Layer Perceptrons (MLPs) $f_{enc}^{\varepsilon}$ and $f_{enc}^{\nu}$ to encode the edge and node features into latent vectors $E_{ij}$ and $V_{i}$, respectively, each of size 64. The encoded graph is represented as $\mathcal{G}_0={(V,E})$.

\textbf{Process}: The processor consists of multiple blocks with unshared weights, where each block performs sequential message passing over the input graph and produces transformed node and edge latent vectors. Residual connections are employed between the input and output edge/node latent vectors to facilitate information flow. The initial block takes the encoded graph as input, and subsequent blocks take the output of the previous block. Within each block, MLPs $f^E$ and $f^V$  are used to apply transformations to the latent edge vector $E_{ij}$ and latent node vector $V_{i}$ using for edges and nodes, respectively.

The edge transformation is described as follows:
$E'{ij}\leftarrow f^E(E{ij}, V_i,V_j)$,

Here, $V_i$ and $V_j$ denote the latent vectors of the sender and receiver nodes, respectively, whereas  $E_{ij}$ represents the latent vector of the connecting edge.

The node transformation is described as follows:
$V'i \leftarrow f^V(V_i,\sum_j{E'{ij}})$

At each node, the transformed latent vectors of incoming edges are aggregated using a permutation-invariant summation function. The resulting sum, along with the node latent vector, is concatenated and fed into the MLP $f^V$. This MLP processes the input and generates the transformed node latent vector, incorporating the information from the aggregated edge vectors.

In Figure \ref{fig:model}, the transformed graph after the first message passing step is denoted as $\mathcal{G}_1={(V', E'})$. The next processor block takes $\mathcal{G}_1$ as input, performs similar transformations with separate MLPs $g^E$ for edges and $g^V$ for nodes, resulting in the transformed graph $\mathcal{G}_2$. This process continues for $M$ message passing blocks, and the output after $M$ blocks, denoted as $\mathcal{G}_M$, serves as input to the decoder.

\textbf{Decoder}: The decoder comprises an edge decoder MLP $f^E_{dec}$ and a node decoder MLP $f^N_{dec}$. This is different from the previous applications of GNNs in dynamics prediction tasks  \cite{sanchez2020learning, pfaff2020} where only the node dynamics are decoded from the node latent vectors using a decoder MLP.
\\
\textbf{Edge Decoder}: The edge decoder MLP takes the latent vectors at each edge as input and predicts a 2D contact force for each edge: $F_{edge}\leftarrow f^E_{dec}(E'_{\mathcal{G}_M})$.\\
\textbf{Node Decoder}: The node decoder MLP takes the latent vectors at each node as input and predicts the net 2D force on each node: $F_{node}\leftarrow f^N_{dec}(V'_{\mathcal{G}_M})$.

%

\section{Case study}
\label{sec:case_study} 
In this study, we utilized the dynamic bearing model described in Section \ref{sec:bearing_model} to simulate trajectories of four bearings with different numbers of rolling elements (13, 14, 15, and 16). These bearings were modeled after the SKF N209 ECP cylindrical roller bearing, which has a pitch diameter of 65.5mm and a roller diameter of 11mm. The length of the rollers is 12mm. Additionally, a horizontal and vertical spring with a stiffness of 5e6N/m is connecting the inner ring to the ground. Dampers with damping ratios of 5e4Ns/m and 1e4Ns/m are used to dampen the inner and outer rings, respectively.
 The simulations were conducted under zero rpm conditions, with an initial external load applied to the outer ring. The range of initial external loads varied from 5000 N to 23000 N, with increments of 2000 N.

\begin{figure}
\centering
\includegraphics[scale=0.5,width=0.5\textwidth]{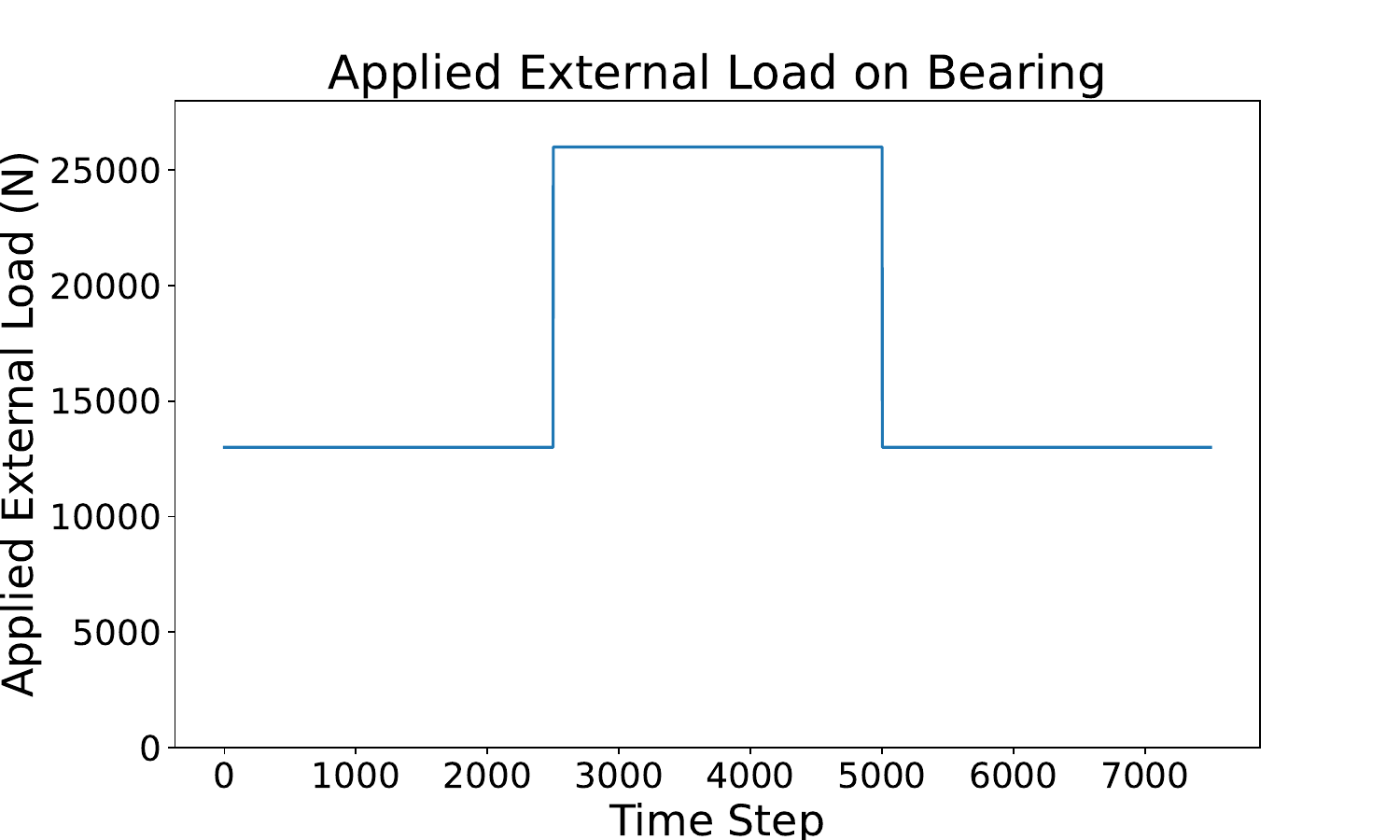}
\caption{Applied external load on the outer ring}
\label{fig:ext_load}
\end{figure}

During each trajectory, an external load was instantaneously applied at the 0th time step. The initial condition of the bearing is all set to 0, so this results in a step response of the system. The external load was doubled at 2500 time steps and subsequently reduced back to the initial load at the 5000th time step. An example of the variation in external load over time is depicted in Figure \ref{fig:ext_load}.

\textbf{Training Data}: The GNN was trained using the trajectories of bearings equipped with 13, 14, and 16 rolling elements. At each time step, the positions and velocities of each component (inner ring, outer ring, and rolling elements) were used to construct the graph representation $\mathcal{G}_t$ of the bearing system. The roller loads were used as the ground truth for the decoded edges, while the total forces acting on each component were used as the ground truth for the decoded nodes. The objective of the GNN was to predict the roller loads and net forces on each component at each time step.

\textbf{Testing Data}: The model was evaluated on the bearing with 15 rollers and an initial load of 13000 N. The applied load during this validation case is shown in figure \ref{fig:ext_load}. We tested its ability to predict the roller loads and net forces on the components given the state vector of each component at a specific time $t$, under different external loads applied to the bearing.

\begin{figure}
\centering
\includegraphics[scale=0.5,width=0.45\textwidth]{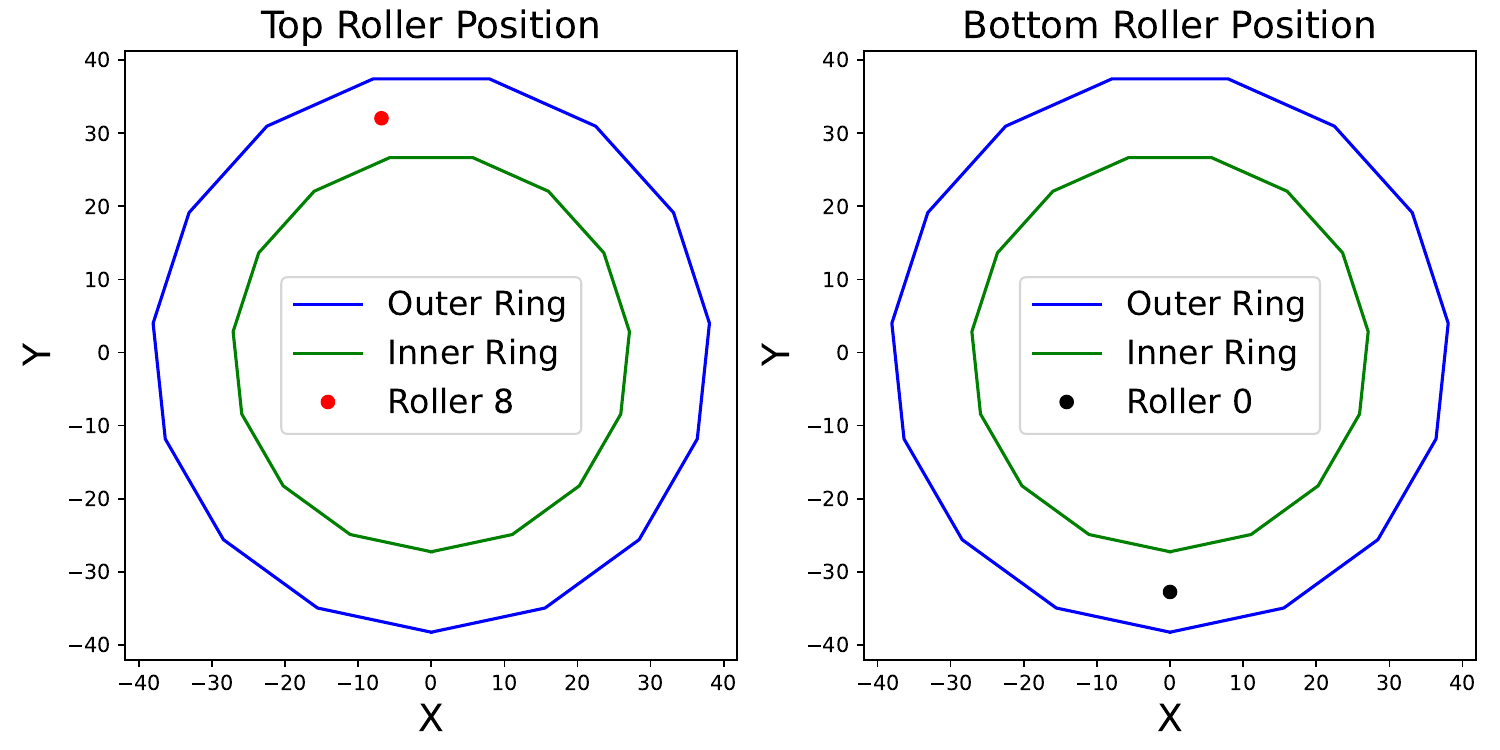}
\caption{Position of top and bottom rolling elements}
\label{fig:top_roller_pos}
\end{figure}

\section{Results}
\label{sec:results} 
\begin{figure}
\centering
\includegraphics[scale=0.8,width=0.45\textwidth]{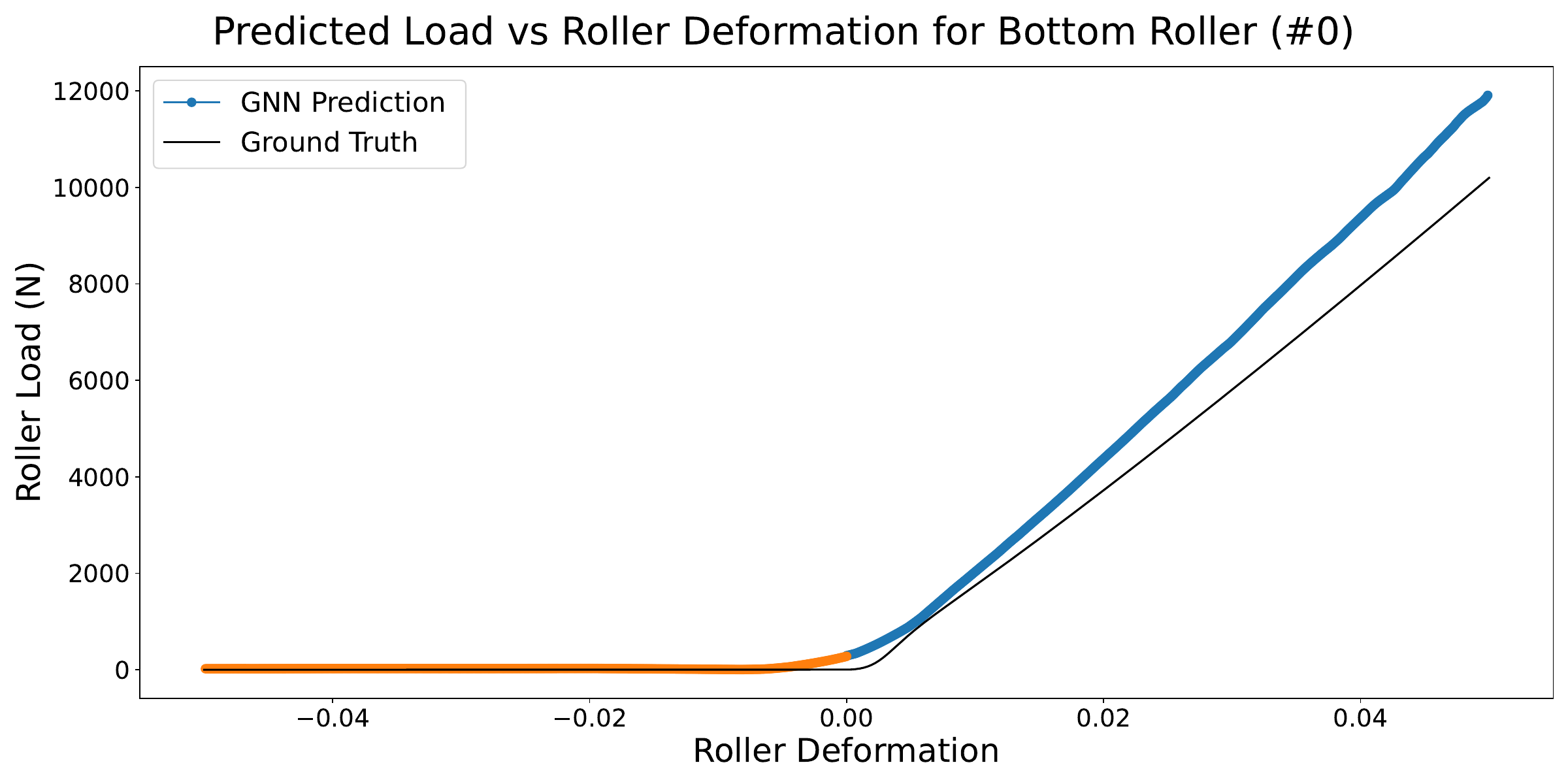}
\caption{Prediction of loading and unloading of bottom roller with inner-ring dispacement}
\label{fig:roller_irdisp}
\end{figure}

In this section, we evaluate the performance of the trained GNN on the test data. We compare the GNN's predictions with those generated by the 2D dynamic model, focusing on a single time step without performing roll-outs. The evaluation covers both the loaded roller (top of the bearing) and the non-loaded roller (bottom of the bearing), as illustrated in Figure \ref{fig:top_roller_pos}. 

\begin{figure*}
\centering
\includegraphics[scale=0.8,width=0.9\textwidth]{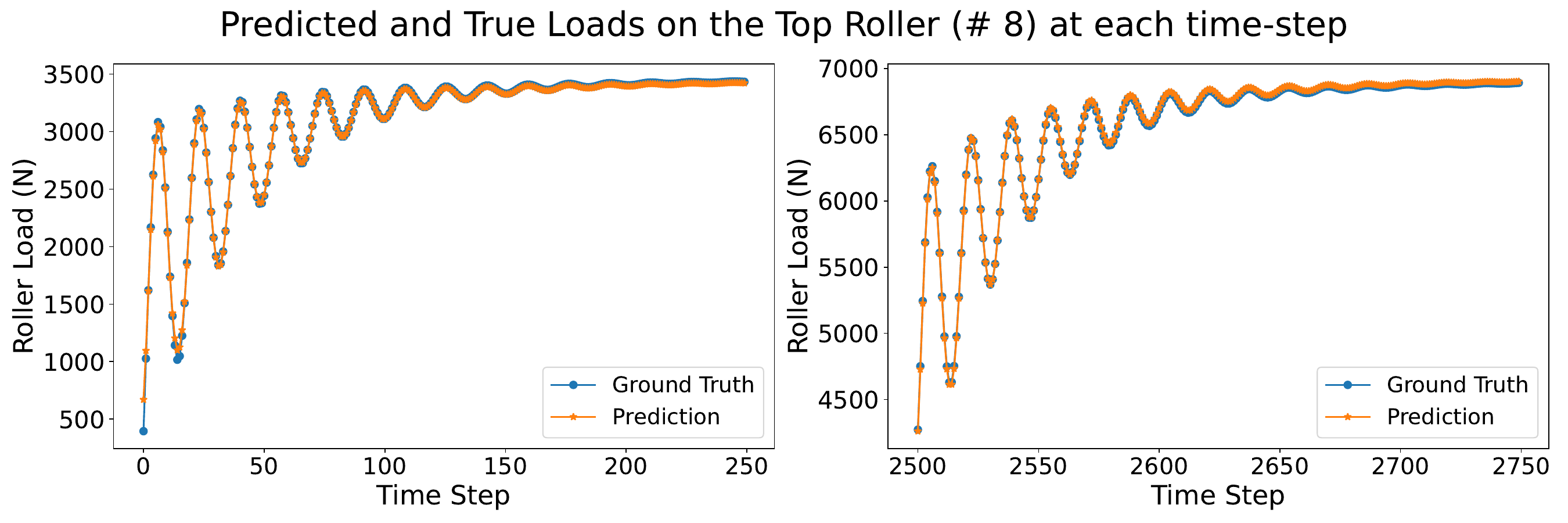}
\caption{Comparison of predictions of roller loads by the GNN for the loaded roller $\# 8$ (shown in Figure \ref{fig:top_roller_pos}) with the results obtained from the dynamic bearing simulator (ground truth). Time-step ranges for plots from left to right: 0-250 $\&$ 2500-2750}
\label{fig:top_roller_load}
\end{figure*}

Figure \ref{fig:top_roller_load} illustrates the predicted loads for the loaded roller number 8 and compares it to the simulated data. It is worth noting the presence of oscillatory dynamics resulting from the sudden application of an external load at the time steps 0 and 2500, as depicted in Figure \ref{fig:ext_load}. These dynamics arise due to the connection of the inner and outer rings to the ground through dampers.

\begin{figure}
\centering
\includegraphics[scale=0.45,width=0.45
\textwidth]{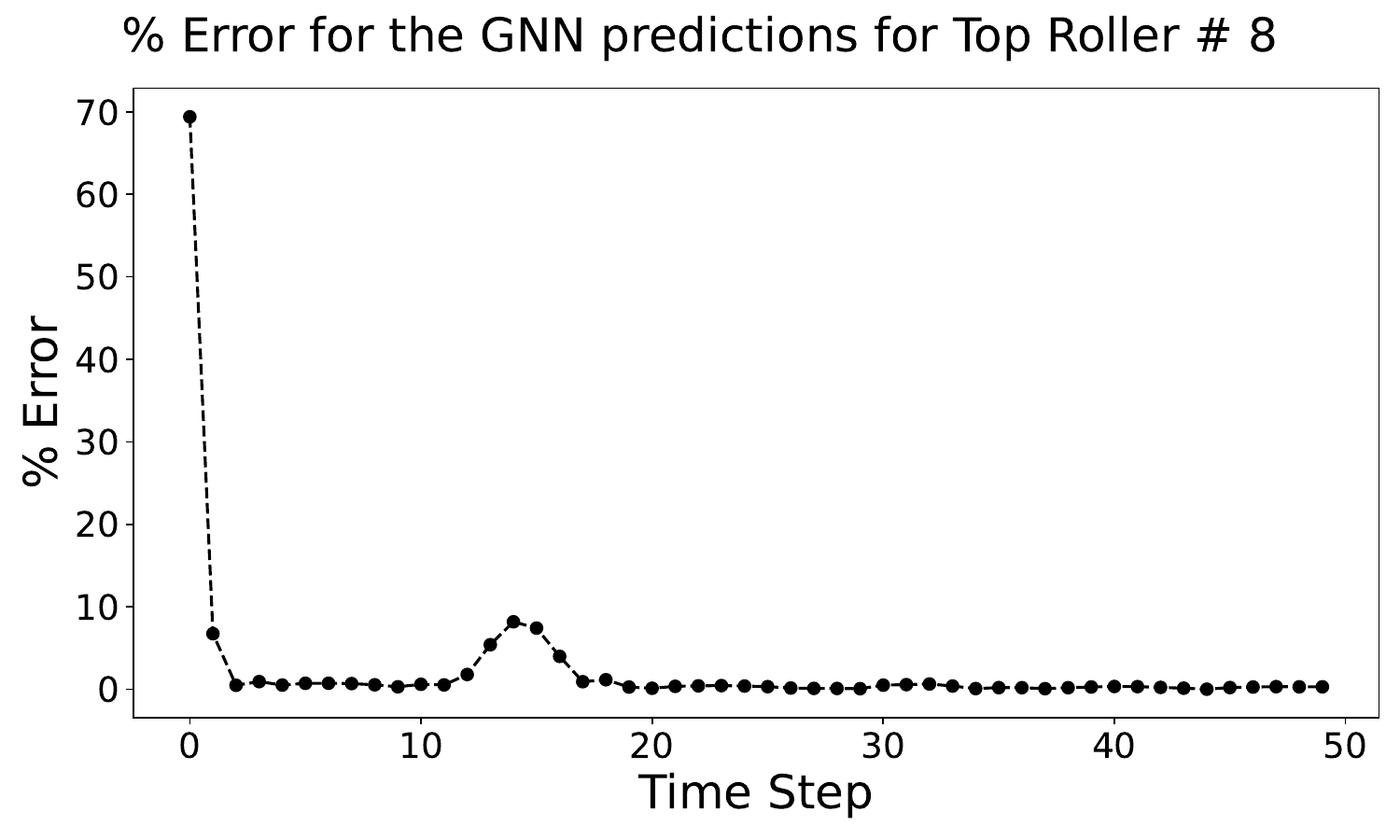}
\caption{Percentage Error in roller load predictions for loaded roller $\# 8$. Error at time step 1 is around 70\%.}
\label{fig:error_top_roller}
\end{figure}

The proposed GNN demonstrates its capability to accurately predict loads even in dynamic regimes of the bearing for the loaded rollers. Moreover, the GNN's performance shows a significant improvement once the bearing reaches a steady state. This improvement is further supported by the percentage error ($\frac{prediction-ground truth}{ground truth}*100\%$) for the loaded rollers, depicted in Figure \ref{fig:error_top_roller} for 50-time steps.

\begin{figure*}
\centering
\includegraphics[scale=0.8,width=0.9
\textwidth]{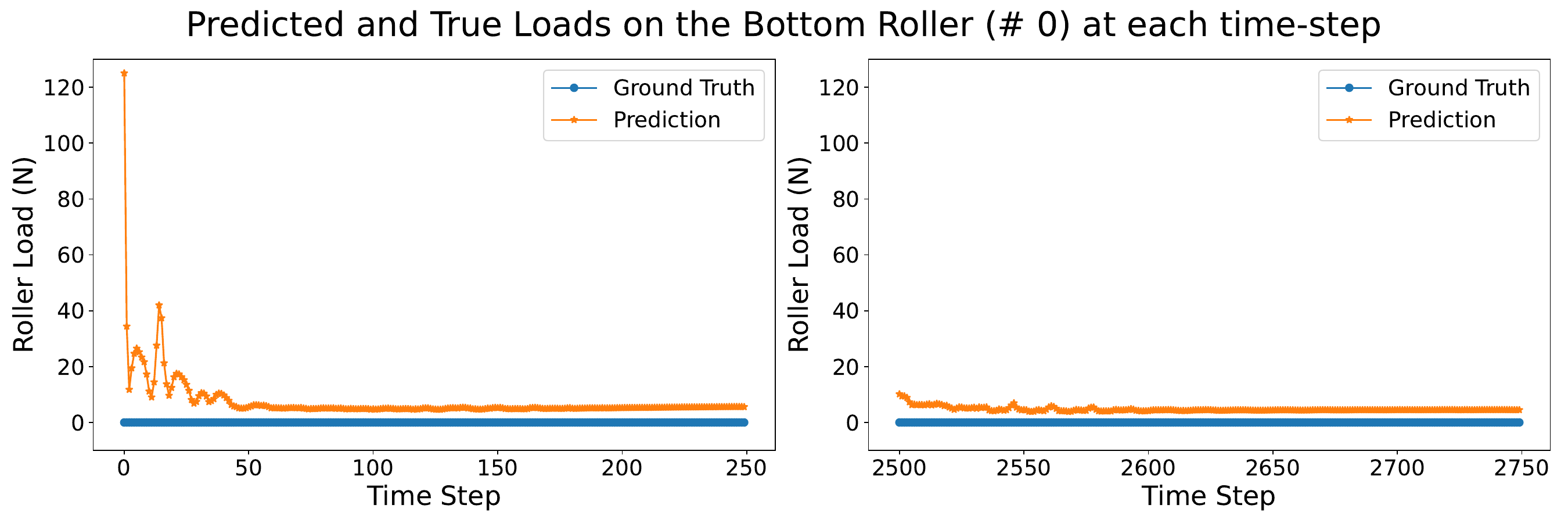}
\caption{Comparison of predictions of roller loads by the GNN for the non-loaded roller $\# 0$ (Figure \ref{fig:top_roller_pos}) with the results obtained from the dynamic bearing simulator (ground truth). Time-step ranges for plots from left to right: 0-250 $\&$ 2500-2750}
\label{fig:bottom_roller_load}
\end{figure*}

Figure \ref{fig:bottom_roller_load} illustrates the predicted load for the unloaded roller. It can be observed that the GNN predicts still small loads for the unloaded rollers even though the ground truth value is zero. While in the first plot in the figure, higher errors in predictions are observed until 50th-time steps, the performance improves once the initial oscillatory dynamics subside. In the second plot, the same observations can be made, however, the magnitude of errors in the initial dynamics phase is lower.

\begin{figure*}
\centering
\includegraphics[scale=0.8,width=0.9
\textwidth]{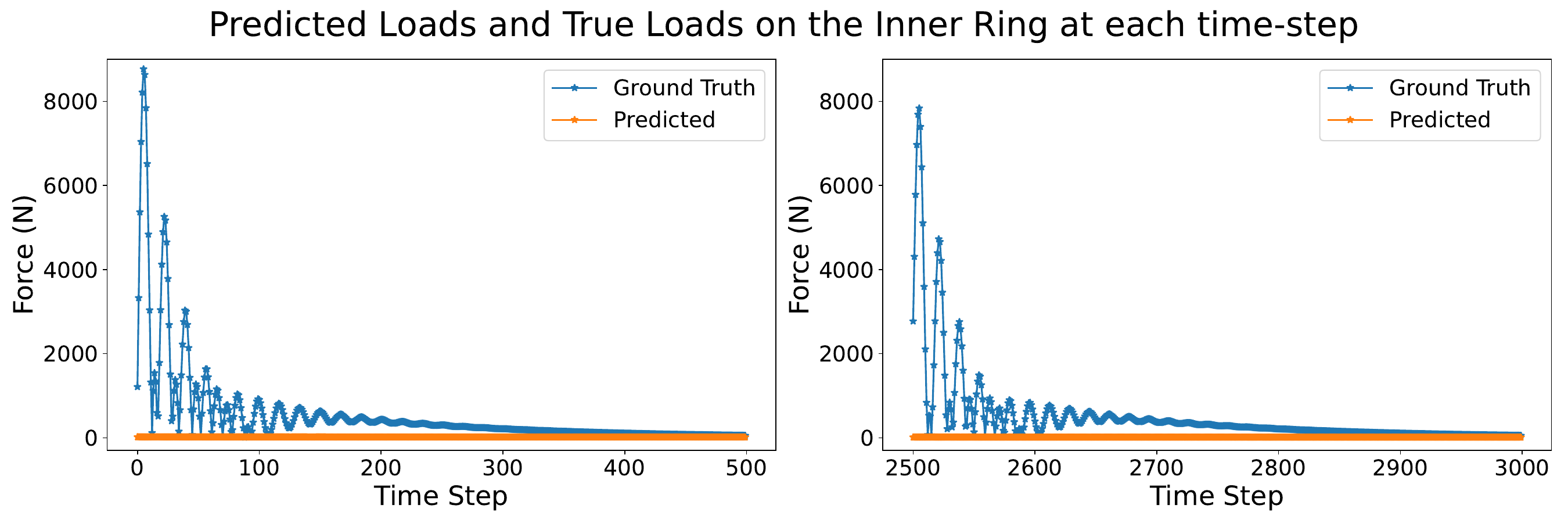}
\caption{Comparison of predicted force on the inner-ring by GNN with the dynamic bearing simulator (ground truth) for single time-step predictions. Time-step ranges for plots from left to right: 0-250 $\&$ 2500-2750}
\label{fig:ir_force}
\end{figure*}

\begin{figure*}
\centering
\includegraphics[scale=0.8,width=0.9
\textwidth]{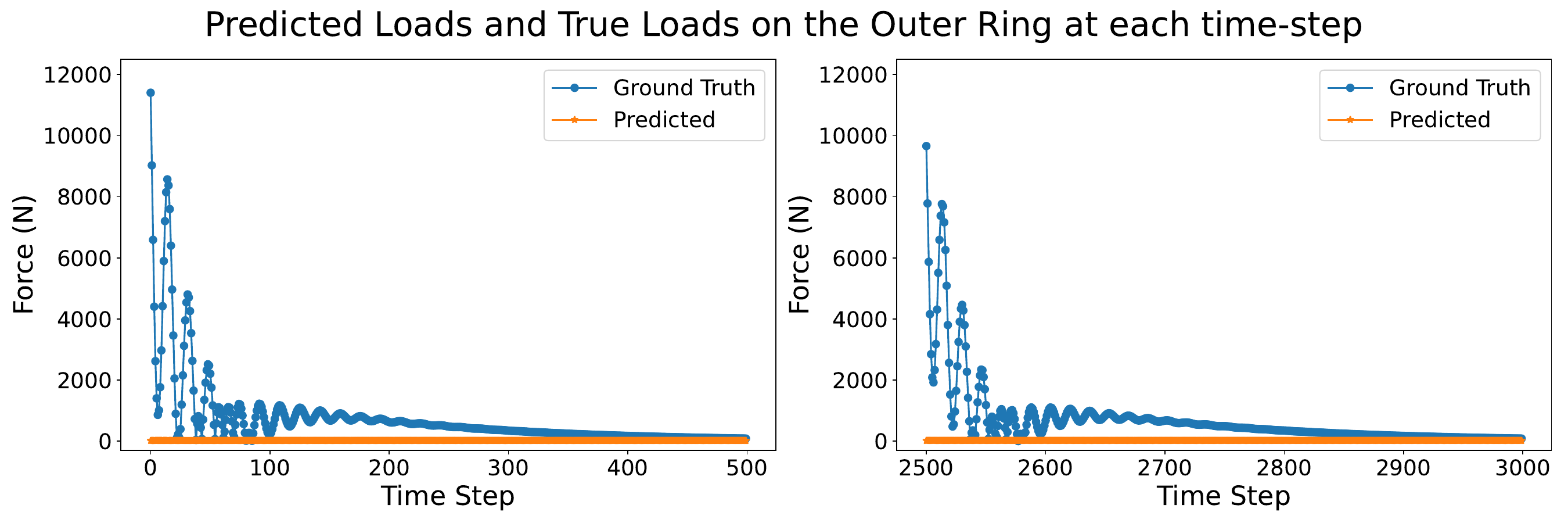}
\caption{Comparison of predicted force on the outer-ring by GNN with the dynamic bearing simulator (ground truth) for single time-step predictions. Time-step ranges for plots from left to right: 0-250 $\&$ 2500-2750}
\label{fig:or_force}
\end{figure*}

Figures \ref{fig:ir_force} and \ref{fig:or_force} present a comparison between the predicted forces on the inner ring and outer ring, respectively, and the ground truth at different time-step ranges. It is evident that when a sudden load is applied at the 0th and 2500th-time steps, both the inner and outer rings experience high dynamical forces. The figures indicate that the GNN predicts a small constant force during these instances, which suggests a limitation in accurately capturing short-term dynamical forces. However, as the rings return to a stable dynamics regime, the GNN demonstrates accurate force predictions

\textbf{Verification of the learned underlying physics}: To verify whether the GNN has learned the correct underlying physics, an artificial trajectory of a bearing with 15 rollers was generated. The experimental setup involved fixing the center of the outer ring at the origin and providing displacement to the inner ring along the y-direction. Initially, the inner ring was centered at the origin and then displaced vertically within the range of -0.05 mm to +0.05 mm. This is equal to a compression of the roller which is located at the bottom-dead-center in the range of -0.05 mm to +0.05 mm.

The generation of the artificial trajectory involved computing the initial positions of each rolling element based on the known radius of the inner and outer rings. We made the assumption that the rolling elements were positioned midway between the circumferences of the inner and outer rings and uniformly distributed along the 360-degree rotation of the bearing.

Figure \ref{fig:roller_irdisp} depicts the predicted and true loads as a function of roller deformation for the bottom dead center roller in the bearing (see Figure \ref{fig:top_roller_pos}). When the inner ring is displaced in the negative y-direction, the bottom-dead-center roller experiences compression, resulting in positive loads. Conversely, displacement of the inner ring in the positive y-direction leads to the unloading of the roller, causing it to experience zero loads. 

The GNN successfully predicts the increase in load for the roller up to a displacement of 0.02 mm of the inner ring in the positive and negative y-direction respectively. Moreover, it is particularly noteworthy that the GNN accurately captures the unloading phenomenon, faithfully reproducing the non-linear loading graph. It can be also noted that the largest deviation between the GNN's prediction and the ground truth is only 15 percent.

These findings demonstrate the GNN's capability to understand and reproduce the expected load changes in response to inner-ring displacements for different rollers within the bearing system.
 This indicates that the GNN has indeed learned the correct underlying physics of the bearing system, as it accurately predicts the expected behavior of the rollers under varying inner-ring displacements.

\section{Conclusions}
\label{sec:conclusions}
This study demonstrates the successful application of a graph neural network framework for predicting the dynamics of bearings. By representing the bearing as a graph and utilizing a message-passing graph neural network, we accurately predict loads at individual time steps based on external load and ring positions/velocities. Our study demonstrates the ability to infer dynamics from trajectory measurements without explicit stiffness, mass, and damping information. In contrast to pure data-driven methods, our approach offers interpretability, generalizability to new conditions such as  external load, and flexibility to adapt to varying bearing configurations.

This proof-of-concept study paves the way for future research, wherein roll-out trajectories can be generated from initial conditions. 
To enhance accuracy, our future research aims to include dampers and springs that connect the rings to the ground. This extension will help address the significant  errors in force predictions on inner and outer rings during the oscillatory dynamics regime that occurs during sudden loading. Additionally, future work will consider including bearing rotation as an important parameter. While our GNN is currently trained on Hertzian contact, it has the potential to capture intricate Elasto-hydrodynamic forces with measured data or FEA simulations, supported by the universal approximation theorem.

This study highlights the potential of graph neural networks in modeling bearing dynamics and opens up new possibilities for advancing bearing diagnostics, prognostics, and the development of real-time operational digital twins for monitoring the health of rotating machinery.

\bibliographystyle{unsrt}  
\bibliography{references}

\begin{thebibliography}{10}

\bibitem{Hag_ehsan}
Ehsan Haghighat, Maziar Raissi, Adrian Moure, Hector Gomez, and Ruben Juanes.
\newblock A deep learning framework for solution and discovery in solid mechanics.
\newblock {\em arXiv: 2003.02751}, 2020.

\bibitem{Yan}
C.A. Yan, R.~Vescovini, and L.~Dozio.
\newblock A framework based on physics-informed neural networks and extreme learning for the analysis of composite structures.
\newblock {\em Computers \& Structures}, 265:106761, 2022.

\bibitem{PINN_alex}
Alexander Henkes, Henning Wessels, and Rolf Mahnken.
\newblock Physics informed neural networks for continuum micromechanics.
\newblock {\em Computer Methods in Applied Mechanics and Engineering}, 393:114790, 2022.

\bibitem{Lutter}
Michael Lutter, Kim Listmann, and Jan Peters.
\newblock Deep lagrangian networks for end-to-end learning of energy-based control for under-actuated systems.
\newblock In {\em 2019 IEEE/RSJ International Conference on Intelligent Robots and Systems (IROS)}, page 7718–7725. IEEE Press, 2019.

\bibitem{greydanus}
Samuel Greydanus, Misko Dzamba, and Jason Yosinski.
\newblock Hamiltonian neural networks.
\newblock {\em Advances in neural information processing systems}, 32, 2019.

\bibitem{jagtapCpinn}
Ameya~D Jagtap, Ehsan Kharazmi, and George~Em Karniadakis.
\newblock Conservative physics-informed neural networks on discrete domains for conservation laws: Applications to forward and inverse problems.
\newblock {\em Computer Methods in Applied Mechanics and Engineering}, 365:113028, 2020.

\bibitem{Gao_PINN}
Han Gao, Matthew~J. Zahr, and Jian-Xun Wang.
\newblock Physics-informed graph neural galerkin networks: A unified framework for solving pde-governed forward and inverse problems.
\newblock {\em Computer Methods in Applied Mechanics and Engineering}, 390:114502, 2022.

\bibitem{battaglia}
Peter Battaglia, Jessica Blake~Chandler Hamrick, Victor Bapst, Alvaro Sanchez, Vinicius Zambaldi, Mateusz Malinowski, Andrea Tacchetti, David Raposo, Adam Santoro, Ryan Faulkner, Caglar Gulcehre, Francis Song, Andy Ballard, Justin Gilmer, George~E. Dahl, Ashish Vaswani, Kelsey Allen, Charles Nash, Victoria~Jayne Langston, Chris Dyer, Nicolas Heess, Daan Wierstra, Pushmeet Kohli, Matt Botvinick, Oriol Vinyals, Yujia Li, and Razvan Pascanu.
\newblock Relational inductive biases, deep learning, and graph networks.
\newblock {\em arXiv: 1806.01261}, 2018.

\bibitem{sanchez2020learning}
Alvaro Sanchez-Gonzalez, Jonathan Godwin, Tobias Pfaff, Rex Ying, Jure Leskovec, and Peter Battaglia.
\newblock Learning to simulate complex physics with graph networks.
\newblock In {\em Proceedings of the International Conference on Machine Learning}, pages 8459--8468, 2020.

\bibitem{pfaff2020}
Tobias Pfaff, Meire Fortunato, Alvaro Sanchez-Gonzalez, and Peter~W Battaglia.
\newblock Learning mesh-based simulation with graph networks.
\newblock In {\em Proceedings of the International Conference on Learning Representations}, 2021.

\bibitem{shlomi2020graph}
Jonathan Shlomi, Peter Battaglia, and Jean-Roch Vlimant.
\newblock Graph neural networks in particle physics.
\newblock {\em Machine Learning: Science and Technology}, 2(2):021001, 2020.

\bibitem{Maosen}
Maosen Li, Siheng Chen, Yangheng Zhao, Ya~Zhang, Yanfeng Wang, and Qi~Tian.
\newblock Dynamic multiscale graph neural networks for 3d skeleton based human motion prediction.
\newblock In {\em Proceedings of the IEEE/CVF conference on computer vision and pattern recognition}, pages 214--223, 2020.

\bibitem{Gupta1984}
Pradeep~K. Gupta, editor.
\newblock {\em Advanced Dynamics of Rolling Elements}.
\newblock Springer New York, 1984.

\bibitem{Lundberg1939}
G.~Lundberg.
\newblock {E}lastische {B}erührung zweier {H}albräume.
\newblock {\em Forschung auf dem Gebiete des Ingenieurwesens}, 10(5):201--211, sep 1939.

\bibitem{palmgren1959}
Arvid Palmgren.
\newblock Ball and roller bearing engineering.
\newblock {\em Philadelphia: SKF Industries Inc}, 1959.

\end{thebibliography}

\end{document}